\documentclass[letterpaper, 10 pt, conference]{ieeeconf}  

\IEEEoverridecommandlockouts                              
\usepackage[utf8]{inputenc}
\usepackage[T1]{fontenc}
\usepackage{graphicx}

\title{\LARGE \bf
Glioblastoma Overall Survival Prediction With Vision Transformers
}

\author{Yin Lin$^{1}$,
        Riccardo Barbieri$^{1}$~\IEEEmembership{Senior Member~IEEE},
        Domenico Aquino$^{1,2}$, 
        Giuseppe Lauria$^{2}$,
        Marina Grisoli$^{2}$, \\
        Elena De Momi$^{1}$~\IEEEmembership{Senior Member~IEEE},
        Alberto Redaelli$^{1}$,
        Simona Ferrante$^{1,2}$~\IEEEmembership{Member~IEEE}
\thanks{$^{1}$ Department of Electronic, Information and Bioengineering, Politecnico di Milano, Milano, Italy}
\thanks{$^{2}$ Department of Neuroradiology, Fondazione IRCCS Istituto Neurologico Carlo Besta, Milano, Italy}
\thanks{$^{1,2}$ Double affiliation, Politecnico di Milano and Fondazione IRCCS Istituto Neurologico Carlo Besta, Milano, Italy}%
}

\begin{document}

\maketitle
\thispagestyle{empty}
\pagestyle{empty}

\begin{abstract}

Glioblastoma is one of the most aggressive and common brain tumors, with a median survival of 10–15 months. Predicting Overall Survival (OS) is critical for personalizing treatment strategies and aligning clinical decisions with patient outcomes. In this study, we propose a novel Artificial Intelligence (AI) approach for OS prediction using Magnetic Resonance Imaging (MRI) images, exploiting Vision Transformers (ViTs) to extract hidden features directly from MRI images, eliminating the need of tumor segmentation. Unlike traditional approaches, our method simplifies the workflow and reduces computational resource requirements.

The proposed model was evaluated on the BRATS dataset, reaching an accuracy of 62.5\% on the test set, comparable to the top-performing methods. Additionally, it demonstrated balanced performance across precision, recall, and F1 score, overcoming the best model in these metrics. The dataset size limits the generalization of the ViT which typically requires larger datasets compared to convolutional neural networks. This limitation in generalization is observed across all the cited studies.
This work highlights the applicability of ViTs for downsampled medical imaging tasks and establishes a foundation for OS prediction models that are computationally efficient and do not rely on segmentation.

\end{abstract}

\section{INTRODUCTION}

Among brain tumors, glioblastoma stands out as one of the most aggressive and common, accounting for 60–70\% of glioma cases \cite{b1,b2}. This tumor is known for its poor prognosis, with a median survival ranging between 10 and 15 months \cite{b1,b3,b4}. In this context, predicting overall survival (OS) in subjects with glioblastoma is crucial, as it allows for the personalization of treatments and the alignment of clinical decisions with individual prognoses.

In recent years, OS prediction has benefited from the application of artificial intelligence learning to magnetic resonance imaging (MRI). Existing studies on OS prediction rely on two primary approaches: a two-phase approach and an approach based on pre-trained models. In the first approach, an initial segmentation model is implemented to identify the tumor in MRI images, then a set of radiomics is extracted on the region of interest (ROI), either manually crafted or generated using deep learning techniques \cite{b14c}. In the subsequent phase, these radiomic features, combined with clinical data, are used to train a classification model. For instance, Chato et. al. \cite{b14b} adopts a simple linear classifier, whereas, in \cite{b5,b6,b7,b8,b9} and \cite{b10}, Random Forest models are exploited. Patel et. al. \cite{b11} utilizes Principal Component Analysis (PCA) to preprocess these features before classification. Meanwhile, \cite{b12, b12a} adopt a Light Gradient-Boosting approach, and \cite{b13,b14,b14a,b14b} apply a neural network for OS prediction. However, this approach requires not only OS labeling but also tumor segmentation, a time-consuming step that clinicians often prefer to avoid. Consequently, the performance of such models is limited by the scarcity of labeled samples, as MRI images are not always accompanied by tumor segmentation. In the second approach, features are extracted using pre-trained models on natural images, which are then used to predict OS \cite{b15}. However, this approach has fundamental limitations due to the structural differences between natural images and MRI scans. Natural images, represented as $n\times m\times 3$ matrices, have three channels (red, green, blue) that share the same spatial structure, with each channel alone being sufficient to recognize objects in the scene. Conversely, in MRI images, each slice represents a distinct plane of a three-dimensional volume, with information that is not directly comparable across slices in the same way as RGB channels. Therefore, using pre-trained models on natural images is inefficient for feature extraction from MRI scans.

A possibile alternative is training a deep learning model that directly uses MRI images as input. However, this strategy faces two main challenges: it requires substantial computational resources, as MRI images have high resolution and large matrix dimensions, and it demands a significant amount of labeled data since the complexity of the input strictly impacts on the number of require samples \cite{b16}. Moreover, several studies have demonstrated that deep learning-based classification models show poorer performance on the BRATS dataset compared to models trained using radiomic features \cite{b16b}.

In recent years, the introduction of transformers in the field of computer vision has opened new possibilities \cite{b17}. Visual transformers, in particular, have demonstrated remarkable success in classifying natural and medical images due to their ability to capture long-range relationships between features \cite{b18}. This paper proposes a novel approach based on the use of visual transformers for OS prediction. MRI images are efficiently preprocessed and used as direct input for a visual transformer model. This method demonstrates competitive performance compared to existing approaches, providing high sensitivity and specificity. Furthermore, it significantly reduces computational requirements, making it a practical and scalable solution for clinical application.

\section{METHODS}

\subsection{Data}
To compare the results with other existing studies, we applied our new procedure to the BRATS dataset \cite{brats1,brats2,brats3}, a widely used dataset for segmentation and prediction of overall survival (OS) in patients with glioblastoma. The dataset includes a total of 494 subjects, of which only 235 have associated OS information. For each subject, data such as age, survival days, and the type of resection performed (gross total/subtotal resection, or no resection) are available. From a medical imaging perspective, the dataset provides four MRI sequences per subject (T1, T1 contrast enhanced, T2, and FLAIR), along with semantic masks for the segmentation of tumor subregions.

Since SubTotal Resection (STR) introduces high variability in outcomes due to the influence of residual tumor, we selected only those subjects who had undergone a gross total resection (GTR), resulting in a total of 118 subjects \cite{b16b}. Each subject is represented by the four MRI sequences, treated as separate samples, for a total of 472 samples. The subjects were divided into a training set (80\%) and a testing set (20\%), ensuring that all sequences from the same subject belonged exclusively to one of the two sets. In the end, the testing set included 96 samples, while the training set contained 376 samples.

The subjects were categorized into three classes, as commonly done in the literature, with thresholds set at 260 and 470 days, corresponding to the $33^{rd}$ and $66^{th}$ percentiles of the OS distribution of the 235 subjects. This choice ensures that all available OS information from the BRATS dataset is incorporated. 


\subsection{Preprocessing}

The MRI images in the BRATS dataset have already undergone standard preprocessing steps, such as deskulling and rigid volume coregistration \cite{brats1}, however, the high resolution of these images (240x240x155) poses a significant computational challenge. To address this issue, we reduced the image resolution to 64x64x50 using third-order spline interpolation. The resulting images were then converted to an 8-bit format, with intensities normalized to the range of 0-255. Comparing to traditional strategies such as radiomic feature extraction or the use of deep features from pre-trained networks, resolution reduction offers a more direct approach. Traditional methods project the image into a lower-dimensional space, providing a condensed representation of the original information. However, this transformation results in the loss of critical details. In contrast, resolution reduction compresses the images using a low-pass filter, which removes higher frequencies while preserving the structural information essential for clinical interpretation. This strategy not only reduces computational complexity but also has the potential to enhance the performance of neural networks by simplifying the input data \cite{b19}.

\subsection{OS\_ViT description}
Building on the basic version of the Visual Transformer (ViT) outlined in \cite{b17}, we applied the ViT directly to our preprocessed MRI images, modifying the patches from 2D to 3D to better match the volumetric nature of the data. Each MRI volume, initially of size $b\times 50\times 64\times 64$ (batch size x depth × height × width), was divided into 640 non-overlapping patches of $b\times 5\times 8\times 8$, each one flattened into a 1-D vector. A 3D convolutional layer is used to project these patches into a 192-dimensional embedding space, producing a feature sequence of shape $b\times 640\times 192$ for each image.

To enable the model to perform better classification, a learnable class token is placed in front of the sequence of patch embeddings. This token acts as an aggregation point for global image features \cite{b17}. Additionally, positional embeddings are added to the patch embeddings and the class token to preserve spatial information within the patch sequence \cite{b17}. The resulting sequence, now of shape $b\times 641\times 192$, is fed into a transformer encoder.

The encoder, based on the architecture described in \cite{b20}, consists of two stacked transformer layers, each featuring multi-head self-attention with 12 heads and a feedforward network (MLP) with an intermediate dimensionality of 1536. Self-attention splits the embeddings into smaller subspaces (16 dimensions per head), computes relationships within and across patches, and recombines them to produce a refined feature representation. The MLP then applies a non-linear transformation to capture higher-order feature interactions.

\begin{figure}
    \centering
    \includegraphics[width=1\linewidth]{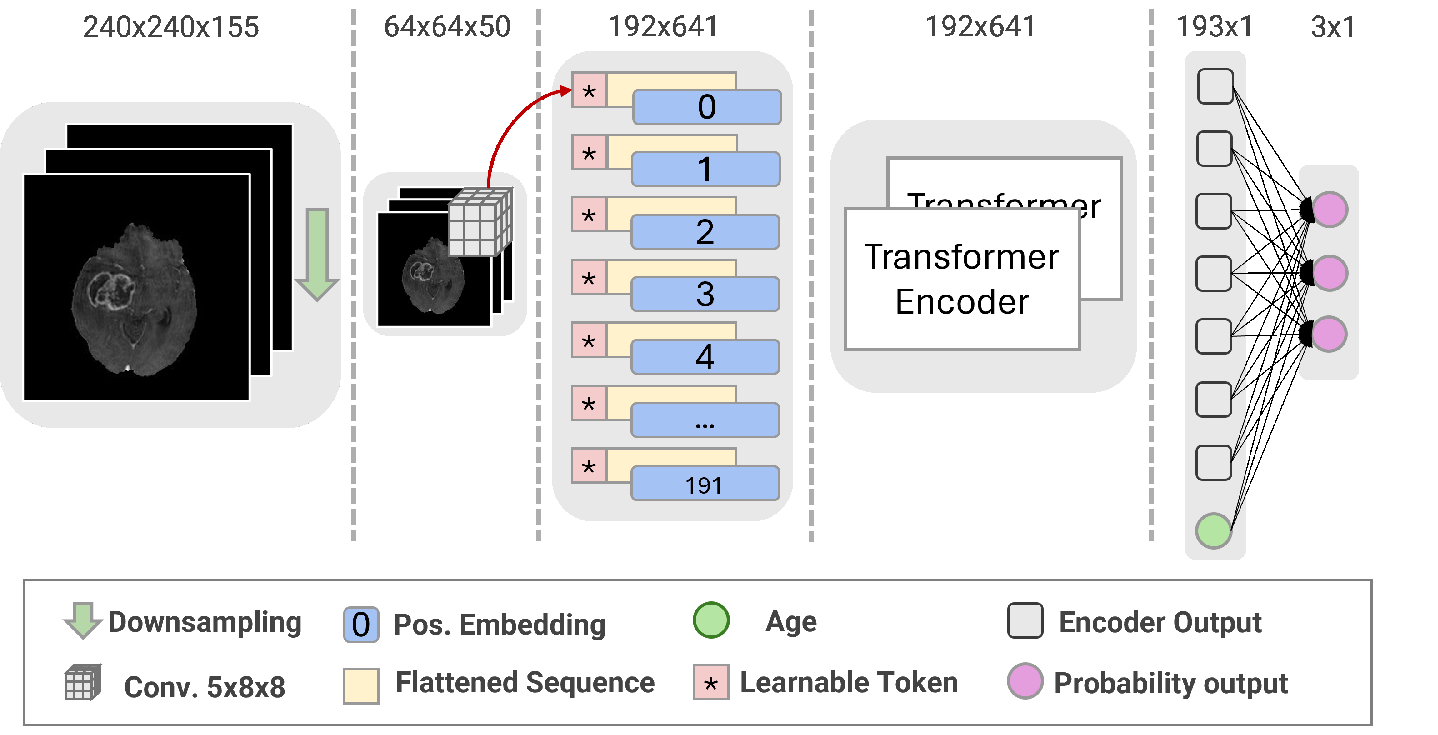}
    \caption{The structure of the OS ViT model is shown. It starts with raw MRI images and follows a structured pipeline to ensure effective feature extraction and classification.}
    \label{process}
\end{figure}

    
After processing by the encoder, the output sequence maintains its original shape (1×641×192). From this sequence, only the class token is retained for the classification task, resulting in a single feature vector of size 1×192. To incorporate clinical information, such as patient age, this feature vector is concatenated with the normalized age value (scaled by 10 to maintain the order of magnitude with the other 192 features), yielding a combined feature vector of size 1×193. Finally, the combined feature vector is passed through a Multi-Layer Perceptron (MLP) head, which includes layer normalization and a fully connected layer that projects the input into three output classes corresponding to short-, medium-, and long-term survival categories. The developed OS\_ViT model, from the raw MRI image input to the transformer’s final output, is reported in Figure \ref{process}.

The hyperparameters of the model were initially configured according to the architecture described in \cite{b17}. They were subsequently fine-tuned manually by analyzing the training loss curves. Once optimized, these hyperparameters were fixed and applied consistently in the test set. However, since all adjustments were made on a single data set, the generalizability of the model can be limited. Future work could overcome this limitation by validating these hyperparameter settings in independent cohorts. The range of model complexity was determined on the basis of the input dimensions, as they are strongly correlated; a highly complex model tends to overfit rapidly. The model was trained with Adam Optimizer with a learning rate set to $10e^{-5}$ and a batch size of 16. The training process utilizes cross-entropy loss as defined in Equation (1, 2). To further avoid the risk of overfitting, an early stopping mechanism was adopted.

\begin{equation}
    \ell(x, y) =\{l_1,\dots,l_N\}^\top
\end{equation}
\begin{equation}
    l_n = \log \frac{\exp(x_{n,y_n})}{\sum_{c=1}^C \exp(x_{n,c})} \cdot 1\{y_n \not= ignore\_index\}
\end{equation}


\section{Results}

\begin{figure}
    \centering
    \includegraphics[width=0.8\linewidth]{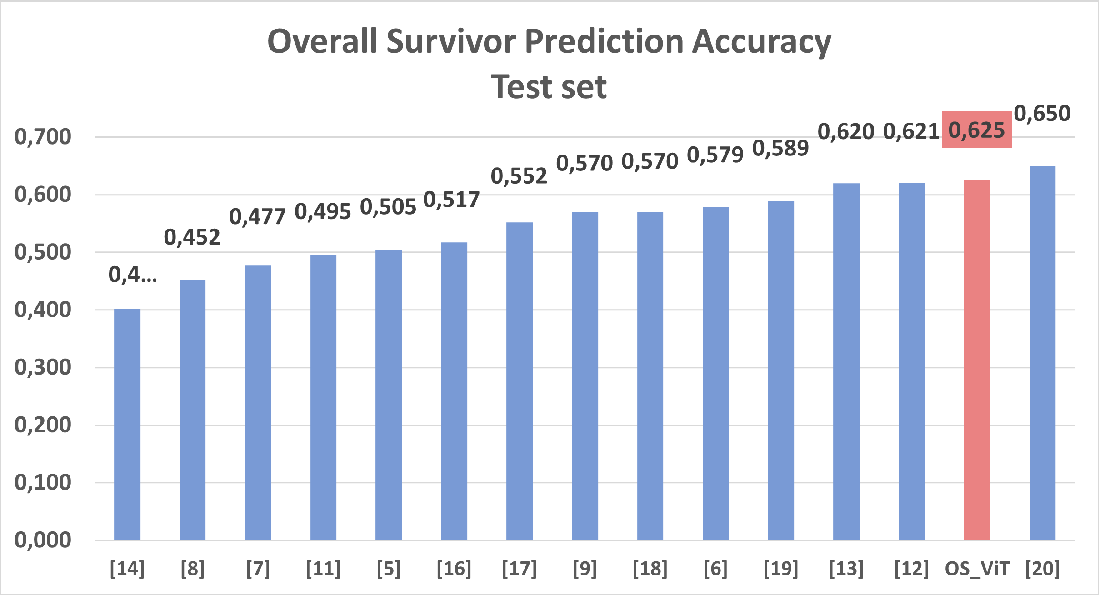}
    \caption{Overview of OS prediction accuracy, including results from 14 existing studies}
    \label{test}
\end{figure}

For each input MRI image, the model outputs a prediction as one of three classes: 0 for long-term survival, 1 for medium-term survival, and 2 for short-term survival. Since most studies report only test set accuracy, comparisons with other approaches are constrained to this metric. Figure \ref{test} offers an overview of the performance across methods, highlighting that our approach achieved second place in classification.

\begin{figure}
    \centering
    \includegraphics[width=0.9\linewidth]{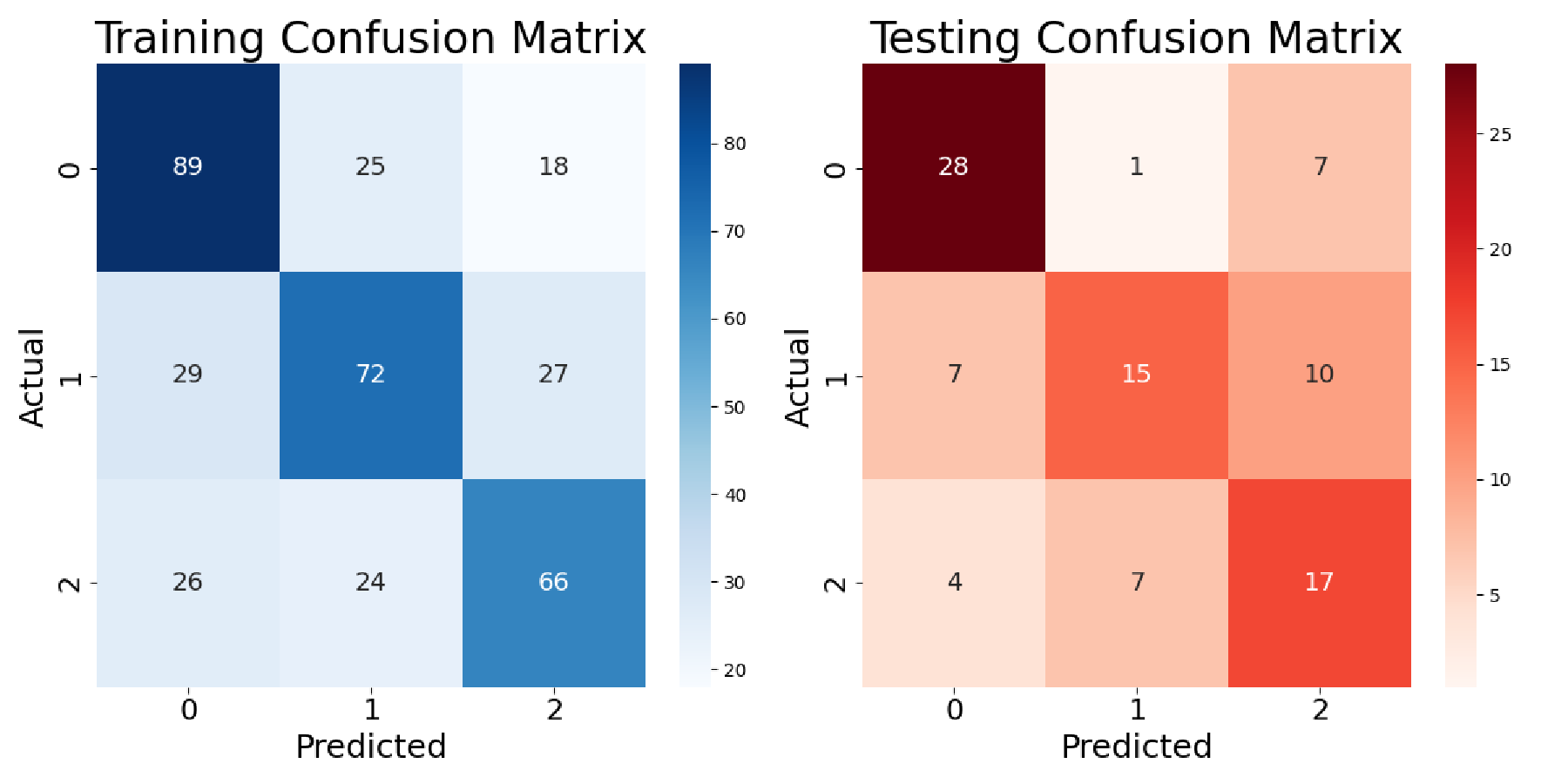}
    \caption{Confusion Matrix: 0 for Long-Term Survivors, 1 for Medium-Term Survivors, and 2 for Short-Term  Survival}
    \label{cm}
\end{figure}

Exploring the specifics of our results, the OS\_ViT model achieved an accuracy of 60.4\%, an average precision of 60.3\%, and an average recall of 60.2\% on the training set, which included 376 samples. On the test set, comprising 96 samples excluded from the training process, the model reached an accuracy of 62.5\%, an average precision of 62.3\%, and an average recall of 61.8\%. The confusion matrices for both the training and testing sets are presented in Figure \ref{cm}.

To provide a more detailed comparison between our model and the one that achieved the highest performance, Table \ref{mt} includes precision, recall, and F1 score for each class as well as their averages. The average metrics are calculated as the arithmetic mean in all classes. As shown in the table, our model achieves a performance comparable to the highest performing approach \cite{b15} in terms of overall precision, recall, and F1 score, surpassing it in two of three classes for each metric. These results highlight the competitive performance of our approach, particularly its balanced behavior across all classes, unlike \cite{b15}, which excels predominantly in long-term survival.

\begin{table}[ht]
    \centering
    \includegraphics[width=0.8\linewidth]{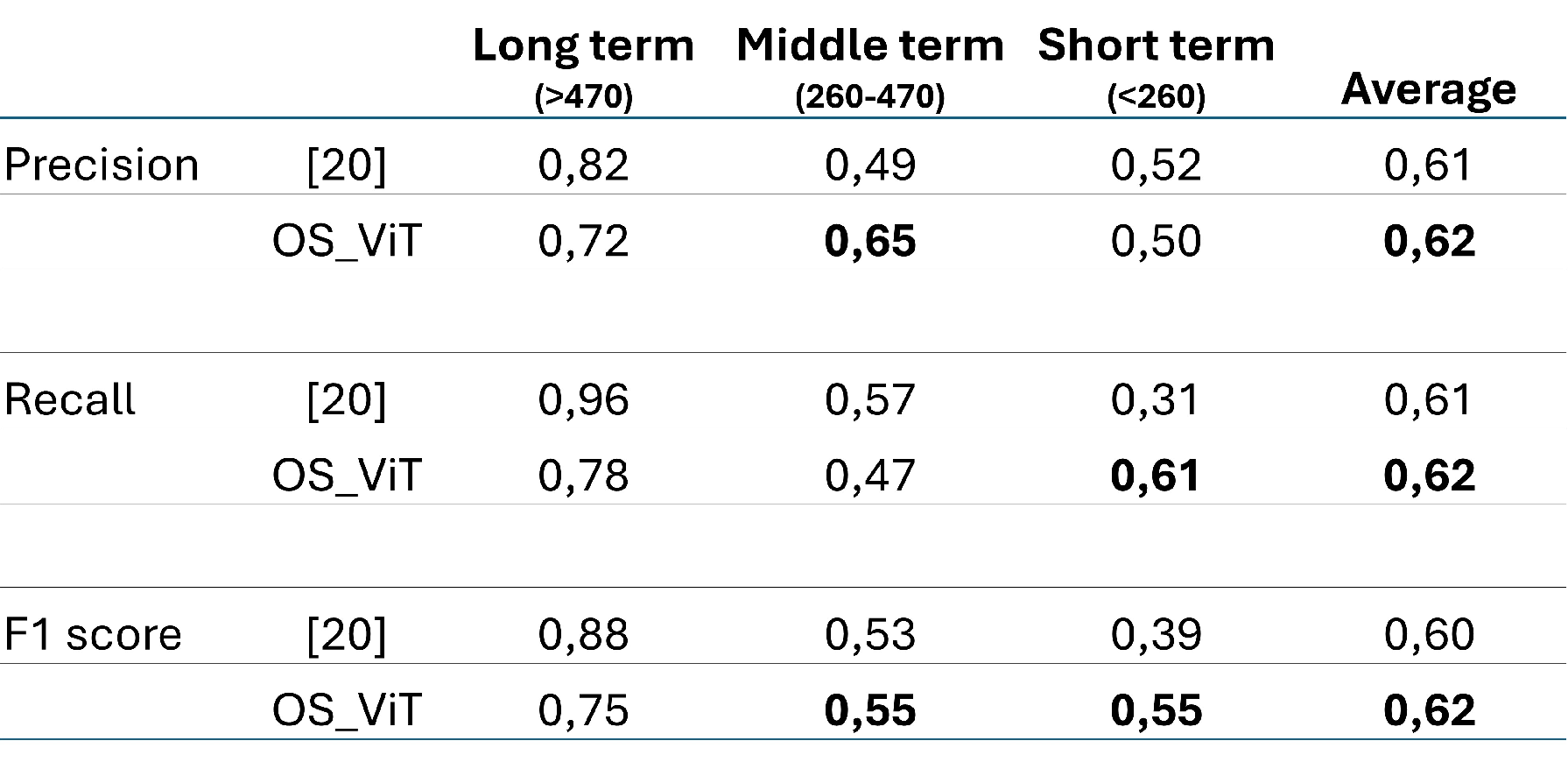}
    \caption{Comparison of Metrics (Precision, Recall, and F1 Score) Between the novel OS\_ViT Model and the Model in [20]}
    \label{mt}
\end{table}

\section{CONCLUSION}

In this study, we proposed a novel approach for OS prediction based on Vision Transformers (OS\_ViT). The proposed OS\_Vit is an architecture capable of directly capturing critical classification features from raw MRI images. The only existing study that does not require segmentation, which achieves the highest accuracy, selects only three slices regardless of tumor characteristics. This approach is clinically questionable, as small tumors may go undetected if they do not appear in the chosen slices. In contrast, our approach eliminates the need for tumor segmentation, significantly simplifying the workflow and reducing computational load through the downsampling of the volume image resolution to 50×64×64. The reduction did not compromise the model's accuracy, confirming that in the context of OS prediction, resolution can be sacrificed while maintaining performance comparable to leading approaches. 

Clinically, OS\_ViT represents a non-invasive and segmentation-free tool solution that can be embedded into current radiology workflows. It operates solely on standard magnetic resonance scans, with minimal preprocessing and no need for tumor delineation, enabling near-real-time inference.

The model achieves comparable results to the existing studies across metrics such as precision, recall, and F1 score. Furthermore, its high precision in classifying patients into several survival groups offers significant benefits for treatment planning. By tailoring therapies to well-defined groups, clinicians can optimize resource allocation, potentially reducing costs as treatments are administered to entire groups rather than on an individual basis. Additionally, the model demonstrates strong predictive performance in the short term and it is particularly valuable from a clinical perspective, since this capability enables the identification of patients who may require more intensive or urgent interventions.

However, a significant limitation of this study is that the model was trained and evaluated solely on the BRATS dataset. This restricts its generalizability and may limit its effectiveness when applied to external clinical data. Vision transformers typically require large-scale datasets to achieve optimal performance, and their effectiveness can degrade when applied to unseen data distributions. Therefore, future work should focus on validating the model using external and multi-institutional datasets to assess robustness and improve clinical applicability.

In conclusion, we successfully applied ViTs to MRI images for OS prediction, achieving results comparable to the best existing studies. Despite some limitations, our study lays the foundation for a segmentation-free OS prediction approach that achieves competitive performance.

\addtolength{\textheight}{-12cm}   




\section*{ACKNOWLEDGMENT}

This work was supported by the project Cal.Hub.Ria (project code T4-AN-09) funded by the Italian Ministry of Health in the framework of  "Piano Sviluppo e Coesione Salute, FSC 2014-2020".


\end{document}